%% file: main.tex
\documentclass[runningheads]{llncs}

\usepackage[T1]{fontenc}
\usepackage{graphicx}
\usepackage{booktabs}
\usepackage{xcolor}
\usepackage{amsmath}
\usepackage{amssymb}

\newenvironment{tightenum}{%
  \begin{list}{\theenumi.}{%
    \usecounter{enumi}%
    \setlength{\topsep}{0pt}%
    \setlength{\partopsep}{0pt}%
    \setlength{\parsep}{0pt}%
    \setlength{\itemsep}{0pt}%
    \setlength{\labelsep}{0.5em}%
    \setlength{\labelwidth}{1.2em}%
    \setlength{\leftmargin}{1.7em}%
    \setlength{\rightmargin}{0pt}%
    \setlength{\listparindent}{0pt}%
  }%
}{\end{list}}

\usepackage{hyperref}

\graphicspath{{./}}

\definecolor{accent}{HTML}{D2691E}
\newcommand{\best}[1]{\textcolor{accent}{\textbf{#1}}}

\newcommand{\trunc}{$\dag$}

\newcommand{\stack}[2]{\shortstack[r]{#1\\[-3pt] \tiny #2}}

\newsavebox{\tblbox}
\newcommand{\fittable}[1]{%
  \sbox{\tblbox}{#1}%
  \ifdim\wd\tblbox>\linewidth
    \resizebox{\linewidth}{!}{\usebox{\tblbox}}%
  \else
    \usebox{\tblbox}%
  \fi}

\begin{document}

\title{Deep Reinforcement Learning for HPC Job Scheduling: A Statistical Evaluation}

\titlerunning{Deep Reinforcement Learning for HPC Job Scheduling}

\author{Justin M. Cheney\inst{1}\orcidID{0009-0004-1991-474X} \and
 Clement N. Nyirenda\inst{1,2}\orcidID{0000-0002-4181-0478}}

\authorrunning{J. M. Cheney, C. N. Nyirenda}

\institute{Computer Science Department, University of the Western Cape,
 Robert Sobukwe Rd, Bellville,
Cape Town, 7535, South Africa\\
Corresponding author: Clement N. Nyirenda (\email{cnyirenda@uwc.ac.za})
\and
eResearch Office, University of the Western Cape,
Bellville, Cape Town, South Africa}

\maketitle
\pagestyle{empty}
\thispagestyle{empty}

\begin{abstract}
Deep Reinforcement Learning (DRL) offers learned job selection for High-Performance Computing (HPC), but performance against heuristics alone does not establish value beyond action masking. Six DRL treatments are compared: Proximal Policy Optimization (PPO), Advantage Actor-Critic (A2C), and Deep Q-Network (DQN), each with and without masking. Three heuristics and a seed-paired masked uniform-random control provide baselines across ten seeds on two real Slurm (Simple Linux Utility for Resource Management) workload traces, one homogeneous and one heterogeneous, under a common non-parametric framework. All masked treatments complete evaluation, whereas unmasked treatments are truncated. On the homogeneous trace, no masked DRL treatment significantly improves average waiting time or average and maximum bounded slowdown over the control. On the heterogeneous trace, MaskablePPO improves all three metrics relative to the control. Its average waiting time is equivalent to the three heuristics within a $\pm10\%$ margin, and its maximum bounded slowdown is 22\% lower than Last-Come, First-Served (LCFS). MaskablePPO has the leading average-waiting rank among complete treatments on both traces, but Nemenyi comparisons detect no significant difference from MaskableA2C. Among masked variants, MaskablePPO has the lowest recorded mean training time, while MaskableDQN has the lowest mean decision latency. These findings distinguish completion from baseline-specific scheduling gains and support including a masked random control when assessing learned-selection value.

\end{abstract}

% ============================================================
% 1  INTRODUCTION
% ============================================================
\section{Introduction}
\label{sec:intro}
High-Performance Computing (HPC) job schedulers determine when jobs run and which resources they receive \cite{Sanjalawe2025}. Increasing workload and hardware diversity complicates the balance between waiting time and resource utilisation \cite{Sanjalawe2025}. Deep Reinforcement Learning (DRL) offers learned scheduling policies as an alternative to traditional heuristics \cite{Firmano2024,Wang2025_1}. Their practical value, however, depends on whether learning improves schedule quality enough to justify training and decision-time costs.

Existing studies benchmark individual DRL schedulers against traditional baselines \cite{Sharma2025,Sankaran2025}, leaving limited evidence for choosing between DRL algorithms under a common experimental protocol. A structured review of 53 papers identifies Proximal Policy Optimization (PPO) family methods in 40\% of the sample, but this prevalence does not establish their superiority. Moreover, performance against a heuristic alone does not establish that learned selection adds value beyond action masking. A controlled comparison must therefore examine algorithm choice and masking together, evaluating schedule quality alongside computational cost.

Action masking excludes infeasible choices before action selection. MaskablePPO, MaskableDQN, and MaskableA2C apply this restriction to their corresponding base algorithms. The objective is to compare DRL algorithms in terms of schedule quality and computational cost, and to assess learning value beyond action masking. The comparison includes six DRL variants: PPO, Deep Q-Network (DQN), and Advantage Actor-Critic (A2C), each with and without masking. Three heuristics and a masked random control provide baselines on two Slurm (Simple Linux Utility for Resource Management) workload traces. Waiting time, slowdown, turnaround time, and resource utilisation are assessed through non-parametric comparisons \cite{Carrasco2020}. The investigation is guided by three research questions (RQs):
\begin{tightenum}
\item What does action masking contribute to schedule quality and evaluation completion?
\item Does PPO's schedule quality and computational cost relative to A2C and DQN justify its prevalence in recent HPC scheduling literature?
\item How does the performance of learned policies relative to heuristic and masked random baselines differ between the homogeneous and heterogeneous workload traces, considering training and decision-time costs?
\end{tightenum}

\noindent This paper makes the following contributions:
\begin{tightenum}
\item A controlled statistical comparison of PPO, A2C, and DQN,
with and without action masking, across homogeneous and heterogeneous
Slurm workload traces, using common workloads, reward functions,
and environment-step budgets.
\item An assessment of masking's contribution to evaluation completion
and schedule quality, including a masked random control to test whether
learned selection adds value beyond feasibility constraints.
\item A non-parametric, multi-metric assessment of algorithm rankings
and trade-offs between schedule quality and cost, evaluating PPO's relative
performance across the two traces.
\end{tightenum}

\noindent Given the paper's scope constraints, a basic understanding of HPC and non-parametric statistical analysis is assumed. The remainder of this paper reviews the literature and the gaps motivating the study (Section~\ref{sec:lit_review}), sets out the methodology (Section~\ref{sec:methodology}), presents the results (Section~\ref{sec:results}), discusses their implications and limitations (Section~\ref{sec:discussion}) and concludes (Section~\ref{sec:conclusion}).

% ============================================================
% 2  LITERATURE REVIEW
% ============================================================
\section{Literature Review}
\label{sec:lit_review}
HPC job scheduling has evolved from rule-based heuristics to learning-based
adaptive techniques. This section reviews that progression and identifies the
methodological gaps motivating the present study.

\subsection{HPC Job Scheduling}
Job scheduling combines job selection, which determines execution order, with resource allocation, which assigns cores, memory, and graphics processing units (GPUs) to selected jobs \cite{Wang2025_1}. Scheduling objectives include reducing waiting times, increasing resource utilisation, and maintaining fairness \cite{Sanjalawe2025}. These objectives can conflict, motivating evaluation across multiple metrics rather than a single performance measure.

\subsection{Traditional and Meta-heuristic Techniques}
Traditional heuristics address competing scheduling objectives by assigning priorities to pending jobs. First-Come-First-Served (FCFS) preserves arrival order, whereas Shortest Job First (SJF) prioritises shorter estimated runtimes. FCFS can cause head-of-line blocking, while SJF can delay long jobs \cite{Silberschatz2018}. Backfilling supplements these priority rules by running other jobs in otherwise idle capacity while protecting a reservation for a waiting job \cite{Lifka1995}. Workload managers such as Slurm coordinate scheduling policies with resource management \cite{Yoo2003}. This separation allows DRL to be studied as an alternative job-selection policy, rather than a replacement for the workload manager. Meta-heuristics search candidate schedules using techniques such as genetic algorithms, particle swarm optimization \cite{subramoney2022multiswarm}, and simulated annealing \cite{Sanjalawe2025}. Such searches can improve schedule quality, but their computational cost and sensitivity to tuning must be considered. Learning-based approaches instead seek a policy that can be reused across scheduling decisions.

\subsection{Deep Reinforcement Learning for Job Scheduling}
Reinforcement learning trains a policy through interaction and reward, without requiring labelled optimal scheduling decisions. DRL uses neural networks to represent policies or value functions, with scheduling formulated as a Markov Decision Process (MDP) over queue and cluster states \cite{Sutton2018,Wang2025_1}. PPO-family methods account for 45\% of the 60 systems described in the surveyed literature. However, performance on general reinforcement-learning benchmarks \cite{Schulman2017} does not by itself establish an advantage in HPC scheduling. This motivates direct comparisons of scheduling performance and computational cost.

\textbf{Algorithm families.} 
Value-based methods estimate action values, with DQN using experience replay and target networks~\cite{mnih2015}. Policy-based methods optimise expected return directly, while actor-critic methods combine a policy with a value estimate that helps reduce variance in its updates. A2C is the synchronous variant of Asynchronous Advantage Actor-Critic (A3C) \cite{mnih2016}. These categories overlap: PPO and A2C both use actor and critic networks, but differ in their update objectives and data reuse. PPO uses a clipped surrogate objective and repeatedly optimises rollout data, whereas A2C uses synchronous advantage-based updates~\cite{mnih2016}. Therefore, their sample efficiency, exploration, and stability must be assessed in the scheduling task rather than inferred from their algorithm category~\cite{bapoo2025parameterized}.

\subsubsection{Action masking.}
Action masking simplifies the agent's decision process by excluding infeasible
actions \cite{Stolz2024}. For value-based methods, invalid actions are masked
before $\arg\max$ selection; for policy-based methods, invalid probabilities are
zeroed before normalisation. In HPC scheduling this prevents selecting jobs
whose resource demands exceed current capacity. When masking is included as
part of the environment without being evaluated separately, reported performance
does not establish whether learning adds value beyond constraint enforcement.
This paper compares masked variants against their unmasked counterparts to
assess differences in schedule quality and evaluation completion. A masked
random control selects uniformly among valid actions, providing a reference
for performance without learning under the same feasibility constraints.

\subsection{DRL Applications in Job Scheduling}
DRL has been adopted across cloud computing \cite{Mangalampalli2024}, edge systems \cite{Wang2024} and HPC clusters \cite{Gao2025,Wang2025_1}, targeting makespan, utilisation and fairness. Wu \emph{et al.} \cite{Wu2025} employed a Double-DQN-derived method for dual-objective flexible job-shop scheduling, comparing it with three meta-heuristics rather than other DRL algorithms. PPO-based examples include Gao \emph{et al.} \cite{Gao2025}, who applied Gated Transformer-XL (GTrXL) with PPO, and Wang \emph{et al.} \cite{Wang2025_1}, who proposed hierarchical PPO for two-level scheduling. These studies evaluate PPO-based scheduling designs rather than providing a common-protocol comparison of PPO, A2C and DQN. RLScheduler \cite{Zhang2020} uses a policy-gradient actor-critic approach, motivated by concerns that workload variability may cause oscillations in Q-learning. Its benchmarks include FCFS, SJF, the named multi-factor priority rules WFP3 and UNICEP (which combine waiting time, requested runtime, and resource demand), and the regression-derived priority rule F1. These heuristic comparisons do not directly compare DRL algorithms or establish learning value against a masked random control. Others extend the base algorithms through multi-agent formulations \cite{Manhary2025} and transfer learning \cite{Wang2025_2}.

These studies demonstrate the potential of DRL for adaptive scheduling, but
differences in statistical treatment, performance metrics, workloads and baselines
limit direct comparison of their reported improvements. A gain against a meta-heuristic on a synthetic workload is not directly equivalent to a gain against a heuristic on a production trace. This evaluation heterogeneity motivates the common-protocol comparison used here.

\subsection{Limitations and Research Gaps}
The benefits of DRL come with limitations bearing on deployability. State and action spaces can grow with workload complexity, increasing the computational demands of learning. Sparse or delayed rewards can complicate learning, particularly when scheduling outcomes emerge over many decisions. Learned policies can be difficult to interpret, which may restrict adoption in mission-critical environments. Algorithm choice therefore involves more than schedule quality. Training requires computation and time, while using a trained policy introduces a decision-time overhead. Interpretability and maintenance present additional deployment concerns, but these are not quantified in this study. Three gaps motivate the present comparison. First, PPO's prevalence does not establish a schedule-quality or computational-cost advantage over A2C and DQN under common experimental conditions. Second, performance with masking alone does not establish learning value beyond feasibility enforcement. Third, differences in metrics, workloads, baselines and statistical treatment limit direct comparison of reported results, while single-metric reporting can conceal trade-offs between schedule quality and utilisation. The methodology in Section~\ref{sec:methodology} addresses these directly.

% ============================================================
% 3  METHODOLOGY
% ============================================================
\section{Methodology}
\label{sec:methodology}
The study compares ten scheduling methods, referred to as treatments: six DRL variants, three heuristics, and one masked random control. PPO, A2C, and DQN use stable-baselines3 (SB3) implementations, while MaskablePPO uses sb3-contrib \cite{stable-baselines3}. MaskableA2C and MaskableDQN use local implementations. The heuristic baselines are SJF, Last-Come, First-Served (LCFS), and UNICEP. All ten are compared across two Slurm workload traces, using the HPCSim scheduling simulator \cite{Wang2025_1} as the environment under fixed seeds for reproducibility. Git commit hashes and SHA-256 manifests record provenance. Locked Nix inputs pin software dependencies, while Apptainer packages the execution environment. Snakemake 9.4.3 orchestrates the pipeline.  

\subsection{Environment and Dataset}
HPCSim uses the cluster and job-queue state representation described by Wang \emph{et al.} \cite{Wang2025_1}. The observation is the information available to the agent at each scheduling decision. The cluster component describes available resources and running jobs. The queue component describes a fixed-size window of pending jobs, including their resource requests, timing information, and scheduling feasibility. The window limits the jobs visible to the agent, not the total queue length. With the configuration in Table~\ref{tbl:environ_config}, queues exceeding 512 jobs are represented by 448 jobs from the head and 64 from the tail. Shorter queues are padded with zeros, and the total queue length is included. The ``best\_fit'' allocator is fixed across all treatments to keep allocation policy constant while comparing job selection. Backfill is disabled in the primary comparisons for both DRL policies and baselines, avoiding an additional scheduling mechanism available only to the heuristics. These comparisons therefore evaluate job-selection policies under fixed allocation and backfill settings, rather than complete production Slurm configurations.

\begin{table}[!b]
\caption{Experimental configuration: environment, hyperparameters and dataset.}
\label{tbl:environ_config}
\label{tbl:hyprparam}
\label{tbl:dataset}
\centering
\fittable{\scriptsize
\begin{tabular}{lll}
\toprule
\textbf{Parameter} & \textbf{Value} & \textbf{Notes} \\
\midrule
\multicolumn{3}{l}{\textit{HPCSim environment}} \\
Allocator & best\_fit & Fixed as controlled factor \\
Window Size & $512$ & Observable queue depth \\
Tail Size & $64$ & Queue tail buffer \\
Random jobs & False & Deterministic trace replay \\
Backfill & False & Uniform mechanism across algorithms \\
\midrule
\multicolumn{3}{l}{\textit{Hyperparameters}} \\
Hidden Layers & $[4096,2048,1024]$ & All \\
Activation function & tanh & All \\
Discount factor ($\gamma$) & $0.99$ & All \\
Policy & MultiInputPolicy & All \\
Entropy Coefficient & 0.0 & PPO/A2C variants only \\
Replay Buffer Size & $150{,}000$ & DQN variants only \\
\midrule
\multicolumn{3}{l}{\textit{Dataset: physical / deeplearn}} \\
Trace & physical\_jobs.csv / deeplearn\_jobs.csv & \\
Topology & physical\_topology.txt / deeplearn\_topology.txt & \\
Node file & nodes.csv & \\
Total Jobs & $84{,}135$ / $68{,}720$ & \\
Development Jobs (70\%) & $58{,}894$ / $48{,}104$ & \\
Holdout Jobs (30\%) & $25{,}241$ / $20{,}616$ & \\
Temporal Range & 37d 11h 42m 8s / 382d 5h 21m 49s & \\
\bottomrule
\end{tabular}}
\end{table}

The dataset is taken from Wang \emph{et al.} \cite{Wang2025_1}: two Slurm traces, one homogeneous (physical), the other heterogeneous (deeplearn). Each trace is ordered by submission time and divided into an earlier 70\% development split and a later 30\% holdout split. The development splits are used for training, primary comparisons, and model selection. The holdout splits are kept for final generalisation assessment.

\subsection{Training and Evaluation}
\label{sec:train-eval}
The DRL variants use the hidden-layer sizes and activation function in Table~\ref{tbl:hyprparam}. Only the DQN variants use a replay buffer. Each DRL training run has a target budget of 3 million environment steps; optimisation-update counts are not matched across algorithms. Ten fixed seeds initialize the Python, NumPy, and PyTorch random number generators. Training six DRL variants with ten seeds produces 60 runs per trace and 120 across both traces. Each trained model is evaluated on its corresponding development split using its training seed. Evaluation is deterministic: the model selects its preferred action rather than sampling exploratory actions. Results are aggregated across the ten seeds for each DRL variant. Scheduling performance is measured using average and maximum waiting time, average and maximum bounded slowdown, average turnaround time, and central processing unit (CPU) and GPU utilisation, following Wang \emph{et al.} \cite{Wang2025_1}. The three primary metrics reported in baseline comparisons are average waiting time and average and maximum bounded slowdown. Decision latency, evaluation wall-clock time, and episode reward are also recorded. The training reward is the negative mean queued waiting time in hours ($R = -\texttt{avg\_waiting} / 3600.0$). This reward is computed every 512 actions; intervening actions receive zero reward.

\subsection{Statistical Framework}
Following Carrasco \emph{et al.} \cite{Carrasco2020}, distributional assumptions are assessed before any comparative tests. Shapiro-Wilk assesses normality of the recorded seed-level metric values for each treatment. The non-normality detected in some treatment distributions, reported in Section~\ref{sec:results}, supports a comparison that does not rely on normally distributed performance values. Friedman therefore compares the six DRL variants through their within-seed ranks across ten seed blocks ($\alpha=0.05$). Kendall's $W = \chi^2_{w}/[N(k-1)]$ accompanies the Friedman results and summarises agreement in treatment ranks across seeds, where $N=10$ and $k=6$. Where Friedman is significant, Nemenyi post-hoc comparisons follow, with Critical Difference (CD) plots showing non-significant rank groupings \cite{Demsar2006}. Wilcoxon confidence intervals and confidence curves give effect-size context. The Page trend test assesses trends across the declared treatment order, not convergence over training time \cite{Derrac2011}. One DRL treatment is Pareto-selected for each trace across the primary and secondary metrics. If needed, the predefined tie-break order is average waiting time, average bounded slowdown, then CPU utilisation. For each heuristic, one-sample Wilcoxon signed-rank tests compare seed-level MaskablePPO and MaskableA2C outcomes with its fixed reference value. Practical equivalence is assessed using two one-sided tests (TOST), with a margin of $\pm$10\% of that reference value \cite{Lakens2017}. Non-significance in a difference test does not establish equivalence. For each DRL treatment compared with a heuristic or the random control, Holm correction is applied jointly to the six Wilcoxon difference-test p-values: average and maximum waiting time, average and maximum bounded slowdown, average turnaround time and CPU utilisation \cite{Holm1979}.

A random action-selection control is included as a reference condition, drawing uniformly at each decision over the valid actions of the same MDP, under the same mask, without backfill, on the same ten seeds. Each masked DRL treatment is compared with the control using a two-sided Wilcoxon signed-rank test on seed-paired differences. The control shows how a policy performs without learning under the same action mask. Matching a heuristic alone does not establish that learning adds value. Finally, DRL treatments and the random control are evaluated on the holdout splits using the same ten seeds; the heuristics remain deterministic reference runs.

% ============================================================
% 4  RESULTS
% ============================================================
\section{Results}
\label{sec:results}

This section reports results following the methodology in Section~\ref{sec:methodology}, presenting the physical and deeplearn traces in turn before examining generalisation, training and inference cost, and training diagnostics. Shapiro-Wilk rejects normality for some average-waiting distributions: physical A2C ($W = 0.634$, $p = 0.00014$) and DQN ($W = 0.753$, $p = 0.0039$), and deeplearn A2C ($p \approx 10^{-7}$). These results support the decision not to assume normality and to use the rank-based framework described in Section~\ref{sec:methodology}. Tests use recorded seed-level metric values; reported means and medians summarise performance across seeds.

Two conventions apply throughout. The first is truncation. None of the three unmasked treatments completed a full evaluation episode on either trace. Without a mask the policy can select an action naming no allocatable job, which HPCSim treats as a request to advance time; once the event queue is empty, simulated time freezes on an identical state and the episode cannot terminate. A hang guard stops the run after 500,000 consecutive steps without a job completion. These metrics describe incomplete evaluations and are not directly comparable with full-trace outcomes. The treatments are excluded from schedule-quality best-value marking and multi-metric selection, and marked ``\trunc'' throughout. They remain in the six-treatment statistical comparisons, which must therefore be interpreted with this limitation. The second is the backfill band. Each heuristic was evaluated with backfill both enabled and disabled, and the width of that band is reported per trace below; the baseline figures quoted elsewhere are the no-backfill ones, for the reason given in Section~\ref{sec:methodology}.

\subsection{Homogeneous Workload: Physical Trace}
\label{sec:phys_results}

On this trace, truncation limits comparisons with the unmasked treatments, while backfill changes the heuristic reference values. Enabling backfill reduces UNICEP's average waiting time by 12.3\%, but LCFS still records lower average waiting time than the complete DRL treatments with both backfill settings. GPU utilisation is omitted because this is a CPU-only workload.

Table~\ref{tbl:phys-main} shows that MaskableA2C and MaskablePPO have similar mean average waiting times, whereas MaskableDQN has higher waiting time and bounded slowdown, with greater variation across seeds. The unmasked variants record higher waiting times still, but their truncated evaluations limit comparisons with complete-trace outcomes. Across all six DRL treatments, Friedman detects statistically significant differences for all nine reported metrics, with $\chi^2$ from 22.11 to 47.14 and Kendall's $W$ from 0.442 to 0.943. However, because these tests include truncated outcomes, the differences do not by themselves establish full-trace schedule-quality superiority. For CPU utilisation, the omnibus test is significant ($p = 2.4 \times 10^{-5}$, $W = 0.579$), yet the mean values for MaskableA2C and MaskablePPO both round to 0.1057; a pairwise comparison is needed to determine whether these two treatments differ.

\begin{table}[!b]
\caption{Physical trace - performance metrics (mean $\pm$ std.). Wait and turnaround in seconds; slowdown is a ratio; utilisation a fraction.}
\label{tbl:phys-main}
\centering
\fittable{\scriptsize\input{tables/physical_main}}
\end{table}

Table~\ref{tbl:phys-rank} ranks MaskablePPO first and MaskableA2C second for average waiting time, but the Nemenyi comparison detects no significant difference between them ($p = 1.000$, $n = 10$, $k = 6$). Across the declared sequence of DQN, A2C, PPO, MaskableDQN, MaskableA2C, and MaskablePPO, the Page test detects a decreasing average-waiting trend ($L = 838$, $p = 6 \times 10^{-6}$), with significant trends also found for the other scheduling metrics in Table~\ref{tbl:phys-main}, except maximum waiting time ($p = 1.0000$). However, these trends depend on the declared order and include truncated evaluations; they do not demonstrate strict monotonic improvement in full-trace schedule quality. For average waiting time on physical, the paired 95\% confidence interval for
MaskableA2C minus MaskablePPO is $[-65.5, +13.3]$\,s. It lies within the stated
$\pm$10\% margin relative to MaskablePPO, supporting practical equivalence for
this metric under that margin. This interval evidence is distinct from the
non-significant Nemenyi comparison and does not establish equivalence across
all metrics. The confidence curve peaks near zero. The recorded multi-metric
selection identifies MaskableA2C as the sole Pareto candidate on physical;
this selection is separate from MaskablePPO's leading average-waiting rank.

\begin{table}[!b]
\caption{Physical trace - Nemenyi post-hoc ranking, average waiting time.}
\label{tbl:phys-rank}
\centering
\fittable{\scriptsize\input{tables/physical_ranks}}
\end{table}

The DRL-only rankings must also be considered alongside performance relative to the heuristics and the masked random control. For MaskablePPO and MaskableA2C, most median relative differences against the no-backfill heuristics lie approximately between $-20\%$ and $+10\%$ on the reported primary metrics. Average bounded slowdown shows larger contrasts: their values are 24.7\% and 24.2\% higher than LCFS, respectively, but 53.8\% and 54.0\% lower than UNICEP.

Table~\ref{tbl:phys-vs-random} compares the three masked DRL treatments with the random control using seed-paired differences. After Holm correction, seven of the nine comparisons detect significant differences, and all seven favour the random control. These comparisons provide no evidence of a learned-policy advantage over masked random selection on the reported metrics under this configuration.

\begin{table}[!b]
\caption{Physical trace - random control comparison, primary metrics.}
\label{tbl:phys-vs-random}
\centering
\fittable{\scriptsize\input{tables/physical_vs_random}}
\end{table}

For mean average waiting time, headroom is defined as the percentage by which
the masked random control exceeds the best no-backfill heuristic, relative to
that heuristic. Table~\ref{tbl:phys-main} gives LCFS at 2,090\,s and the random
control at 2,166\,s; the unrounded values yield 3.7\% headroom. On this physical
trace, the results do not establish a learned-policy advantage over the masked
random control that would justify replacing a heuristic.

\subsection{Heterogeneous Workload: Deeplearn Trace}
\label{sec:dl_results}

Table~\ref{tbl:dl-main} shows that MaskablePPO's mean average waiting time is close to that of the no-backfill heuristics. Enabling backfill reduces LCFS's average waiting time by 5.7\%, while SJF and UNICEP then record means about 4 and 3 seconds lower than MaskablePPO, respectively. The baseline gap is larger on this trace. Using mean average waiting time, Table~\ref{tbl:dl-main} gives the masked random control at 1,170\,s and the best no-backfill heuristic, UNICEP, at 846\,s. The unrounded values yield 38.4\% headroom relative to UNICEP, compared with 3.7\% on physical. This descriptive gap does not establish why learning performs differently across traces.  Across the six DRL treatments, Friedman detects statistically significant differences for all ten metrics, with Kendall's $W$ from 0.487 to 0.905. However, GPU utilisation ($\chi^2 = 28.63$, $W = 0.573$) is computed as $\text{resource-seconds}/$\allowbreak$(\text{total resource} \times \text{makespan})$; for fixed completed work, a longer makespan lowers the reported ratio. This measure should therefore be interpreted alongside makespan and evaluation completion. Evaluation wall-clock time for unmasked A2C, DQN and PPO also includes the 500,000 stall steps before truncation.

\begin{table}[!b]
\caption{Deeplearn trace - performance metrics (mean $\pm$ std. dev.). Wait and turnaround in seconds; slowdown is a ratio; utilisation a fraction.}
\label{tbl:dl-main}
\centering
\fittable{\scriptsize\input{tables/deeplearn_main}}
\end{table}

Table~\ref{tbl:dl-ranks} gives unmasked PPO the lowest mean rank for average waiting time, but its truncated evaluations make it ineligible for selection. MaskablePPO has the lowest mean rank among complete treatments and is the recorded multi-metric selection. However, no Nemenyi comparison with MaskablePPO detects a significant difference. MaskableA2C's lower median (837\,s versus 851\,s) contrasts with its much higher mean, which is dominated by two extreme seed-level outcomes. The lower median alone therefore does not establish superiority or consistent performance across seeds.

\begin{table}[!b]
\caption{Deeplearn trace - Nemenyi post-hoc ranking, average waiting time.}
\label{tbl:dl-ranks}
\centering
\fittable{\scriptsize\input{tables/deeplearn_ranks}}
\end{table}

Table~\ref{tbl:dl-vs-random} shows statistically significant improvements for MaskablePPO over the masked random control on all three reported primary metrics after Holm correction. Against the no-backfill heuristics, however, TOST supports practical equivalence with UNICEP on all three primary metrics within the $\pm$10\% margin. The reported reductions in maximum bounded slowdown are 22.0\% relative to LCFS and 63.2\% relative to the random control.

\begin{table}[!b]
\caption{Deeplearn trace - random control comparison, primary metrics.}
\label{tbl:dl-vs-random}
\centering
\fittable{\scriptsize\input{tables/deeplearn_vs_random}}
\end{table}

Table~\ref{tbl:dl-main} shows that MaskablePPO's mean average waiting time is within 1\% of SJF (851\,s) and UNICEP (846\,s). These small average-waiting differences should be considered alongside the training and decision-time costs in Section~\ref{sec:train_inf_cost}. The configured deeplearn topology contains 30 nodes and 120 GPUs,
while physical contains 87 nodes and 6,264 CPU cores. Differences between these
two traces do not establish that heterogeneous resources cause better DRL
performance.

\subsection{Generalisation}
Generalisation is assessed on the reserved holdout splits without further training or reselection. On physical, mean average waiting time falls approximately fourfold for MaskableA2C and MaskablePPO; on deeplearn, it rises approximately threefold for MaskablePPO. These changes concern different chronological job sets and should not alone be interpreted as improvement or deterioration in the trained policies. The ranks order treatments by their across-seed mean average waiting times, rather than the mean within-seed ranks used in the DRL statistical comparisons. MaskableA2C and MaskablePPO retain first and second place on physical, while MaskablePPO and unmasked PPO exchange the first two positions on deeplearn. Elsewhere on physical, MaskableDQN moves from third to sixth and A2C from fourth to third.

\subsection{Training and Inference Cost}
\label{sec:train_inf_cost}
Tables~\ref{tbl:phys-cost} and~\ref{tbl:dl-cost} show that, among the masked variants, MaskablePPO and MaskableA2C have lower mean training times than MaskableDQN on both traces, while MaskableDQN has the lowest mean decision latency. Focusing on these variants keeps the latency comparison within completed evaluations.

\begin{table}[!b]
\caption{Physical trace - training and inference cost.}
\label{tbl:phys-cost}
\centering
\fittable{\scriptsize\input{tables/physical_cost}}
\end{table}

\begin{table}[!b]
\caption{Deeplearn trace - training and inference cost.}
\label{tbl:dl-cost}
\centering
\fittable{\scriptsize\input{tables/deeplearn_cost}}
\end{table}

MaskableDQN's lower decision latency in Tables~\ref{tbl:phys-cost} and~\ref{tbl:dl-cost} is only one part of its resource cost: training also requires replay storage. The reported comparisons do not isolate the effects of replay capacity or update reuse on schedule quality, but the training configuration provides context for the diagnostics in Section~\ref{sec:training_diagnostics}. Both DQN variants use a buffer capacity of 150,000 transitions and minibatches of 2,048, storing current and next observations from the high-dimensional queue and cluster state. The memory requirements of this storage depend on array dimensions and data types.

\subsection{Training Diagnostics}
\label{sec:training_diagnostics}
Explained variance, entropy, and learning rate are examined to help interpret the scheduling results. These diagnostics do not establish that training was sufficient or isolate the cause of poor performance. No episode-reward curve is reported because \texttt{rollout/ep\_rew\_mean} changes infrequently during training. DQN logs reward at episode boundaries, with reported averages of 3 to 21 reward-log entries per run. Its blank explained-variance and entropy entries reflect metrics not reported by the logger, rather than zero-valued measurements. 

The training diagnostics first examine policy entropy, which describes the concentration of action probabilities. For unmasked A2C, the final logged entropy is $0.07 \pm 0.16$ nats on physical and $6.23 \pm 0.00$ on deeplearn. The former indicates concentrated action probabilities, whereas the latter is close to the 513-action uniform-distribution ceiling ($\ln 513 \approx 6.24$). Critic fit is examined through explained variance, which measures how much variation in training returns is captured by the critic, rather than schedule quality: 1 indicates no residual variation and 0 indicates no reduction in variance. For example, A2C's windowed explained variance on the physical trace is $0.944 \pm 0.018$ over the 2M--3M training-step interval.

The distinction between critic fit and schedule quality can be examined through the masked treatments. On physical, the final logged explained variance is $0.533 \pm 0.451$ for MaskableA2C and $0.109 \pm 0.263$ for MaskablePPO. Neither treatment differs significantly from the masked random control in average waiting time, while both have significantly higher average and maximum slowdown (Table~\ref{tbl:phys-vs-random}). These scheduling comparisons, rather than critic fit alone, provide the evidence for assessing whether learned job selection adds value beyond masking.

% ============================================================
\section{Discussion}
\label{sec:discussion}
Tables~\ref{tbl:phys-main} and~\ref{tbl:dl-main} highlight the importance of headroom. The descriptive values of 3.7\% on physical and 38.4\% on deeplearn affect how each workload's results are interpreted, without establishing heterogeneity as their cause. The training diagnostics also highlight two issues: the common 3M environment-step target does not match optimiser update counts, and rewards occur only every 512th action. At $\gamma = 0.99$, a reward 512 actions away carries a discount weight of $0.99^{512} \approx 0.0058$. Controlled comparisons of reward density and optimisation budgets within each trace would help resolve their contributions to scheduling outcomes. The findings remain specific to the tested reward, training configurations and two workload traces. These constraints frame the research questions below.

\subsubsection{RQ1: The contribution of action masking.}

All masked treatments complete evaluation on both traces, but completion alone does not establish good schedule quality. The masked random comparisons assess the additional value of learning. On physical, neither MaskablePPO nor MaskableA2C differs significantly from the control in average waiting time, while both have significantly higher average and maximum slowdown (Table~\ref{tbl:phys-vs-random}). On deeplearn, MaskablePPO improves all three metrics relative to the control, with Holm-adjusted $p = 0.012$ for each comparison (Table~\ref{tbl:dl-vs-random}). MaskableDQN performs significantly worse than the control on all three metrics on both traces. Masking therefore supports completion in the tested configuration, while the benefit of learned selection depends on the treatment and workload.

\subsubsection{RQ2: The justification for concentrating on PPO.}

MaskablePPO has the best average-waiting rank among complete treatments on both traces (Tables~\ref{tbl:phys-rank} and~\ref{tbl:dl-ranks}), but its Nemenyi comparisons with MaskableA2C detect no significant difference on physical ($p = 1.000$, rounded) or deeplearn ($p = 0.999$, rounded). The recorded multi-metric selection chooses MaskableA2C on physical and MaskablePPO on deeplearn. The cost comparison also presents a trade-off: PPO variants have lower recorded mean training times than their corresponding A2C variants (by 9--22\%) and DQN variants, while MaskableDQN has the lowest mean decision latency among masked treatments (Tables~\ref{tbl:phys-cost} and~\ref{tbl:dl-cost}). These findings support PPO as a leading candidate in the tested configuration, but do not establish a general schedule-quality and cost advantage that justifies its prevalence.

\subsubsection{RQ3: When a DRL scheduler is worth its inference cost.}

With heuristic backfill disabled, the justification for learned selection differs between the two traces. On physical, the masked treatments show no improvement over the random control in average waiting time or slowdown. On deeplearn, MaskablePPO has significantly lower average slowdown than all three heuristics and is equivalent to UNICEP on the three primary metrics within the specified $\pm10\%$ margin. These gains carry computational costs: MaskablePPO's evaluation takes 35$\times$ and 13$\times$ the fastest heuristic's time on physical and deeplearn, respectively. The deeplearn holdout further qualifies the comparison: its descriptive average-waiting advantage over LCFS and SJF widens, but its difference from UNICEP remains below 1\%, while its advantage over masked random narrows from 27.7\% to 3.9\%. Practical justification therefore depends on whether gains over the relevant baseline, on the metrics of interest, outweigh training and decision-time costs under the intended workload.

% ============================================================
% 7  CONCLUSION
% ============================================================
\section{Conclusion and Future Work}
\label{sec:conclusion}
The study compared six DRL treatments, three heuristics and a masked uniform-random control over ten seeds on two real HPC workload traces under a common non-parametric framework. Masking contributed to the completion of evaluation runs, but the findings also show that the benefit of learned selection depends on both the treatment and workload, rather than masking alone. MaskablePPO had the leading average-waiting rank among complete treatments, but its Nemenyi comparisons with MaskableA2C detected no significant difference. The multi-metric selection chose MaskableA2C on physical and MaskablePPO on deeplearn. PPO's lower recorded training times and DQN's lower decision latencies present a cost trade-off rather than a general advantage for one family. Scheduling gains were also baseline- and metric-dependent: on the deeplearn holdout, MaskablePPO's average-waiting difference from UNICEP remained below 1\%. Practical justification therefore requires weighing the relevant gains against training and decision-time costs.

Beyond these findings, a key outcome is the value of including a seed-paired masked random control. It helps distinguish the performance of learned selection from that attainable through feasibility masking with random selection. Building on this comparison, future studies should vary reward density and optimisation budgets within fixed traces to assess their contributions to scheduling outcomes, then examine additional workloads to test whether the observed contrasts recur. If consistent learned-selection benefits are established, model distillation could be evaluated for reducing decision latency while preserving feasibility masking and schedule quality.

\begin{credits}
\subsubsection{\ackname}
Computations were performed using the University of the Western Cape's High Performance Computing and Research Cloud facilities. The authors acknowledge the use of OpenCode to assist with grammatical editing, sentence restructuring, and minor code refinement. All suggestions assisted by artificial intelligence were reviewed, revised, and verified by the authors, who take full responsibility for the final content and code.
\subsubsection{\discintname}
The authors have no competing interests to declare that are relevant to the content of this article.
\end{credits}

% ---- Bibliography ----
\label{endofbody}
\clearpage
\bibliographystyle{splncs04}
\bibliography{./bibliography}

\end{document}

%% file: tables/physical_main.tex
\begin{tabular*}{\linewidth}{@{\extracolsep{\fill}}lrrrrrr@{}}
\toprule
\textbf{Algorithm} & \stack{\textbf{Avg}}{\textbf{Wait}} & \stack{\textbf{Avg}}{\textbf{Slowdown}} & \stack{\textbf{Avg}}{\textbf{Turnaround}} & \stack{\textbf{Max}}{\textbf{Wait}} & \stack{\textbf{Max}}{\textbf{Slowdown}} & \stack{\textbf{CPU}}{\textbf{Util}} \\
\midrule
MaskablePPO & \stack{2,243}{$\pm$41} & \stack{7.49}{$\pm$0.34} & \stack{10,985}{$\pm$41} & \stack{69,126}{$\pm$18,255} & \stack{2,918}{$\pm$44} & \stack{0.1057}{$\pm$0.0000} \\
MaskableDQN & \stack{4,474}{$\pm$1,684} & \stack{116.39}{$\pm$85.42} & \stack{13,216}{$\pm$1,684} & \stack{2,065,995}{$\pm$1,309,114} & \stack{193,701}{$\pm$122,640} & \stack{0.0982}{$\pm$0.0070} \\
MaskableA2C & \stack{2,223}{$\pm$42} & \stack{7.36}{$\pm$0.14} & \stack{10,965}{$\pm$42} & \best{\stack{49,881}{$\pm$17,225}} & \stack{2,966}{$\pm$67} & \stack{0.1057}{$\pm$0.0000} \\
PPO$\dag$ & \stack{8,470}{$\pm$6,177} & \stack{125.33}{$\pm$120.52} & \stack{17,065}{$\pm$6,159} & \stack{47,693}{$\pm$13,031} & \stack{4,060}{$\pm$949} & \stack{0.0974}{$\pm$0.0029} \\
DQN$\dag$ & \stack{6,229}{$\pm$4,478} & \stack{87.23}{$\pm$84.90} & \stack{14,820}{$\pm$4,495} & \stack{54,745}{$\pm$40,343} & \stack{5,325}{$\pm$4,034} & \stack{0.0980}{$\pm$0.0027} \\
A2C$\dag$ & \stack{5,500}{$\pm$4,816} & \stack{63.99}{$\pm$92.35} & \stack{14,094}{$\pm$4,803} & \stack{34,767}{$\pm$6,897} & \stack{3,357}{$\pm$531} & \stack{0.0968}{$\pm$0.0006} \\
\midrule
LCFS & \best{2,090} & \best{5.88} & \best{10,832} & 58,418 & 3,096 & \best{0.1060} \\
SJF & 2,176 & 8.14 & 10,918 & 69,909 & 3,060 & 0.1051 \\
UNICEP & 2,499 & 15.87 & 11,241 & 63,851 & 3,513 & 0.1053 \\
Random (masked) & \stack{2,166}{$\pm$76} & \stack{6.01}{$\pm$0.19} & \stack{10,908}{$\pm$76} & \stack{53,819}{$\pm$18,028} & \best{\stack{2,599}{$\pm$175}} & \stack{0.1058}{$\pm$0.0001} \\
\bottomrule
\end{tabular*}%

%% file: tables/physical_ranks.tex
\begin{tabular*}{\linewidth}{@{\extracolsep{\fill}}lrrrr@{}}
\toprule
\textbf{Algorithm} & \textbf{Mean Rank} & \textbf{Median} & \textbf{Mean} & \textbf{p vs. best} \\
\midrule
MaskablePPO & \best{1.7} & 2,242 & 2,243 & --- \\
MaskableA2C & 1.9 & 2,212 & 2,223 & 1.000 \\
A2C$\dag$ & 4.1 & 2,923 & 5,500 & 0.047* \\
MaskableDQN & 4.4 & 5,008 & 4,474 & 0.016* \\
PPO$\dag$ & 4.4 & 8,623 & 8,470 & 0.016* \\
DQN$\dag$ & 4.5 & 4,429 & 6,229 & 0.011* \\
\bottomrule
\end{tabular*}%

%% file: tables/physical_vs_random.tex
\begin{tabular*}{\linewidth}{@{\extracolsep{\fill}}llrrrcrl@{}}
\toprule
\textbf{Treatment} & \textbf{Metric} & \stack{\textbf{Random}}{\textbf{median}} & \textbf{DRL median} & \textbf{$\Delta$\%} & \stack{\textbf{Seeds}}{\textbf{better}} & \textbf{p (Holm)} & \textbf{vs. random} \\
\midrule
MaskablePPO & Avg Wait & 2,171.033 & 2,242.491 & +4.3\% & 2/10 & 0.156 & Not sep. \\
MaskablePPO & Avg Slow. & 5.977 & 7.335 & +24.8\% & 0/10 & 0.012 & Worse \\
MaskablePPO & Max Slow. & 2,570.333 & 2,914.091 & +12.9\% & 1/10 & 0.029 & Worse \\
MaskableDQN & Avg Wait & 2,171.033 & 5,007.956 & +139.9\% & 1/10 & 0.012 & Worse \\
MaskableDQN & Avg Slow. & 5.977 & 125.576 & +1982.5\% & 0/10 & 0.012 & Worse \\
MaskableDQN & Max Slow. & 2,570.333 & 251,871.750 & +9392.2\% & 0/10 & 0.012 & Worse \\
MaskableA2C & Avg Wait & 2,171.033 & 2,212.443 & +3.1\% & 3/10 & 0.336 & Not sep. \\
MaskableA2C & Avg Slow. & 5.977 & 7.305 & +22.7\% & 0/10 & 0.012 & Worse \\
MaskableA2C & Max Slow. & 2,570.333 & 2,962.182 & +14.6\% & 0/10 & 0.012 & Worse \\
\bottomrule
\end{tabular*}%

%% file: tables/deeplearn_main.tex
\begin{tabular*}{\linewidth}{@{\extracolsep{\fill}}lrrrrrrr@{}}
\toprule
\textbf{Algorithm} & \stack{\textbf{Avg}}{\textbf{Wait}} & \stack{\textbf{Avg}}{\textbf{Slowdown}} & \stack{\textbf{Avg}}{\textbf{Turnaround}} & \stack{\textbf{Max}}{\textbf{Wait}} & \stack{\textbf{Max}}{\textbf{Slowdown}} & \stack{\textbf{CPU}}{\textbf{Util}} & \stack{\textbf{GPU}}{\textbf{Util}} \\
\midrule
MaskablePPO & \stack{846}{$\pm$9} & \best{\stack{3.64}{$\pm$0.03}} & \stack{11,413}{$\pm$9} & \best{\stack{31,442}{$\pm$153}} & \best{\stack{2,261}{$\pm$58}} & \best{\stack{0.1172}{$\pm$0.0000}} & \best{\stack{0.2612}{$\pm$0.0000}} \\
MaskableDQN & \stack{45,729}{$\pm$40,335} & \stack{853.29}{$\pm$810.96} & \stack{56,296}{$\pm$40,335} & \stack{19,003,961}{$\pm$12,150,302} & \stack{1,182,525}{$\pm$1,035,356} & \stack{0.1095}{$\pm$0.0055} & \stack{0.2440}{$\pm$0.0123} \\
MaskableA2C & \stack{24,520,863}{$\pm$77,476,701} & \stack{670,447.13}{$\pm$2,118,679.67} & \stack{24,531,430}{$\pm$77,476,701} & \stack{51,687,957}{$\pm$161,407,893} & \stack{5,128,953}{$\pm$16,058,567} & \stack{0.1060}{$\pm$0.0349} & \stack{0.2361}{$\pm$0.0778} \\
PPO$\dag$ & \stack{818}{$\pm$9} & \stack{3.66}{$\pm$0.01} & \stack{11,241}{$\pm$25} & \stack{30,684}{$\pm$0} & \stack{2,291}{$\pm$0} & \stack{0.1158}{$\pm$0.0002} & \stack{0.2559}{$\pm$0.0009} \\
DQN$\dag$ & \stack{19,328}{$\pm$51,265} & \stack{446.45}{$\pm$1,125.33} & \stack{29,736}{$\pm$51,262} & \stack{5,263,952}{$\pm$10,301,905} & \stack{284,036}{$\pm$761,632} & \stack{0.1152}{$\pm$0.0017} & \stack{0.2542}{$\pm$0.0036} \\
A2C$\dag$ & \stack{4,274}{$\pm$10,923} & \stack{66.55}{$\pm$198.87} & \stack{14,706}{$\pm$10,970} & \stack{521,773}{$\pm$1,552,986} & \stack{41,571}{$\pm$123,972} & \stack{0.1154}{$\pm$0.0009} & \stack{0.2550}{$\pm$0.0014} \\
\midrule
LCFS & 912 & 3.75 & 11,479 & 32,083 & 2,907 & \best{0.1172} & \best{0.2612} \\
SJF & 851 & 3.95 & 11,418 & 31,513 & 2,291 & \best{0.1172} & \best{0.2612} \\
UNICEP & \best{846} & 3.76 & \best{11,413} & 31,513 & 2,291 & \best{0.1172} & \best{0.2612} \\
Random (masked) & \stack{1,170}{$\pm$26} & \stack{7.23}{$\pm$0.29} & \stack{11,737}{$\pm$26} & \stack{678,917}{$\pm$237,272} & \stack{7,004}{$\pm$3,476} & \stack{0.1172}{$\pm$0.0000} & \stack{0.2612}{$\pm$0.0000} \\
\bottomrule
\end{tabular*}%

%% file: tables/deeplearn_ranks.tex
\begin{tabular*}{\linewidth}{@{\extracolsep{\fill}}lrrrr@{}}
\toprule
\textbf{Algorithm} & \textbf{Mean Rank} & \textbf{Median} & \textbf{Mean} & \textbf{p vs. best} \\
\midrule
PPO$\dag$ & 1.7 & 821 & 818 & 0.066 \\
A2C$\dag$ & 2.2 & 821 & 4,274 & 0.261 \\
DQN$\dag$ & 3.1 & 825 & 19,328 & 0.891 \\
MaskablePPO & \best{4.0} & 851 & 846 & --- \\
MaskableA2C & 4.3 & 837 & 24,520,863 & 0.999 \\
MaskableDQN & 5.7 & 45,552 & 45,729 & 0.324 \\
\bottomrule
\end{tabular*}%

%% file: tables/deeplearn_vs_random.tex
\begin{tabular*}{\linewidth}{@{\extracolsep{\fill}}llrrrcrl@{}}
\toprule
\textbf{Treatment} & \textbf{Metric} & \stack{\textbf{Random}}{\textbf{median}} & \textbf{DRL median} & \textbf{$\Delta$\%} & \stack{\textbf{Seeds}}{\textbf{better}} & \textbf{p (Holm)} & \textbf{vs. random} \\
\midrule
MaskablePPO & Avg Wait & 1,171.719 & 850.901 & -28.0\% & 10/10 & 0.012 & Better \\
MaskablePPO & Avg Slow. & 7.199 & 3.645 & -49.8\% & 10/10 & 0.012 & Better \\
MaskablePPO & Max Slow. & 5,921.342 & 2,267.550 & -63.2\% & 10/10 & 0.012 & Better \\
MaskableDQN & Avg Wait & 1,171.719 & 45,551.950 & +3900.3\% & 1/10 & 0.023 & Worse \\
MaskableDQN & Avg Slow. & 7.199 & 721.107 & +9995.8\% & 1/10 & 0.023 & Worse \\
MaskableDQN & Max Slow. & 5,921.342 & 957,643.358 & +12070.5\% & 1/10 & 0.023 & Worse \\
MaskableA2C & Avg Wait & 1,171.719 & 837.357 & -27.8\% & 8/10 & 1.000 & Not sep. \\
MaskableA2C & Avg Slow. & 7.199 & 3.646 & -48.6\% & 8/10 & 1.000 & Not sep. \\
MaskableA2C & Max Slow. & 5,921.342 & 2,307.000 & -55.7\% & 8/10 & 1.000 & Not sep. \\
\bottomrule
\end{tabular*}%

%% file: tables/physical_cost.tex
\begin{tabular*}{\linewidth}{@{\extracolsep{\fill}}lrcrrr@{}}
\toprule
\textbf{Algorithm} & \textbf{Train (h)} & \textbf{Runs} & \textbf{Eval (s)} & \textbf{Eval Ratio} & \stack{\textbf{Latency}}{\textbf{(ms)}} \\
\midrule
MaskablePPO & 3.81 $\pm$ 0.09 & 10 & 2,677 & 35$\times$ & 22.31 \\
MaskableDQN & 5.76 $\pm$ 0.68 & 10 & \best{1,434} & 19$\times$ & 2.77 \\
MaskableA2C & 4.34 $\pm$ 0.17 & 10 & 2,612 & 34$\times$ & 22.55 \\
PPO$\dag$ & \best{3.42 $\pm$ 0.24} & 10 & 19,842 & 257$\times$ & 23.91 \\
DQN$\dag$ & 5.63 $\pm$ 0.66 & 10 & 6,297 & 81$\times$ & \best{2.68} \\
A2C$\dag$ & 3.77 $\pm$ 0.12 & 10 & 18,954 & 245$\times$ & 20.94 \\
\bottomrule
\end{tabular*}%

%% file: tables/deeplearn_cost.tex
\begin{tabular*}{\linewidth}{@{\extracolsep{\fill}}lrcrrr@{}}
\toprule
\textbf{Algorithm} & \textbf{Train (h)} & \textbf{Runs} & \textbf{Eval (s)} & \textbf{Eval Ratio} & \stack{\textbf{Latency}}{\textbf{(ms)}} \\
\midrule
MaskablePPO & \best{1.23 $\pm$ 0.02} & 10 & 629 & 13$\times$ & 4.97 \\
MaskableDQN & 1.73 $\pm$ 0.02 & 10 & \best{352} & 7$\times$ & \best{0.67} \\
MaskableA2C & 1.45 $\pm$ 0.27 & 10 & 1,202 & 25$\times$ & 5.06 \\
PPO$\dag$ & 1.26 $\pm$ 0.25 & 10 & 4,542 & 94$\times$ & 4.74 \\
DQN$\dag$ & 1.71 $\pm$ 0.02 & 10 & 2,509 & 52$\times$ & 1.08 \\
A2C$\dag$ & 1.61 $\pm$ 0.29 & 10 & 5,022 & 104$\times$ & 4.89 \\
\bottomrule
\end{tabular*}%